\documentclass[letterpaper, 10 pt, conference]{ieeeconf}  % Comment this line out if you need a4paper

\IEEEoverridecommandlockouts                              % This command is only needed if 
\usepackage{graphicx} % for pdf, bitmapped graphics files
\usepackage{amsmath} % assumes amsmath package installed
\usepackage{caption}
\usepackage{subcaption}
\usepackage{hyperref}
\usepackage{amssymb}  % assumes amsmath package installed

\title{\LARGE \bf
Temp-Frustum Net: 3D Object Detection with Temporal Fusion
}

\author{Eme\c{c} Er\c{c}elik$^{1}$, Ekim Yurtsever$^{2}$, and Alois Knoll$^{1}$% <-this % stops a space
\thanks{$^{1}$Eme\c{c} Er\c{c}elik and Alois Knoll are with the Chair of Robotics, Artificial Intelligence and Real-time Systems,
        Technical University of Munich, 85748 Garching b. M\"{u}nchen, Germany
        {\tt\small emec.ercelik@tum.de}, % 
        {\tt\small knoll@in.tum.de}}%
\thanks{$^{2}$Ekim Yurtsever is with the Ohio State University, Columbus, OH 43212, USA
        {\tt\small yurtsever.2@osu.edu}}%
}

\begin{document}

\maketitle
\thispagestyle{empty}
\pagestyle{empty}

%%%%%%%%%%%%%%%%%%%%%%%%%%%%%%%%%%%%%%%%%%%%%%%%%%%%%%%%%%%%%%%%%%%%%%%%%%%%%%%%
\begin{abstract}
3D object detection is a core component of automated driving systems. State-of-the-art methods fuse RGB imagery and LiDAR point cloud data frame-by-frame for 3D bounding box regression. However, frame-by-frame 3D object detection suffers from noise, field-of-view obstruction, and sparsity. We propose a novel Temporal Fusion Module (TFM) to use information from previous time-steps to mitigate these problems. First, a state-of-the-art frustum network extracts point cloud features from raw RGB and LiDAR point cloud data frame-by-frame. Then, our TFM module fuses these features with a recurrent neural network. As a result, 3D object detection becomes robust against single frame failures and transient occlusions. Experiments on the KITTI object tracking dataset show the efficiency of the proposed TFM, where we obtain ~6\%, ~4\%, and ~6\% improvements on Car, Pedestrian, and Cyclist classes, respectively, compared to frame-by-frame baselines. Furthermore, ablation studies reinforce that the subject of improvement is temporal fusion and show the effects of different placements of TFM in the object detection pipeline. Our code is open-source and available at \url{https://github.com/emecercelik/Temp-Frustum-Net.git}.

\end{abstract}

\section{INTRODUCTION}

Traffic participants are mostly very dynamic and some of them are not visible to the ego-vehicle's sensors for a time period due to temporal occlusions. To fully observe the environment and to make the detection and planning easier, it is very beneficial to detect traffic participants all the time,  which can be fully- or partly-occluded to the ego-vehicle in different time intervals. Therefore, the detection information of an object in previous time-steps becomes essential, which is also obtained for free, and can be used to compensate the loss of information due to especially occlusions in the current time-step for a better detection. 

According to the well-known 3D object detection benchmarks KITTI \cite{geiger2012} and nuScenes\cite{caesar2020}, the best performing 3D object detection methods make use of single time-step of sensor data (LiDAR). However, this means ignoring the valuable detection and feature information already calculated in the previous time-steps. 

\begin{figure}[t]
    \centering
    \includegraphics[width=0.47\textwidth]{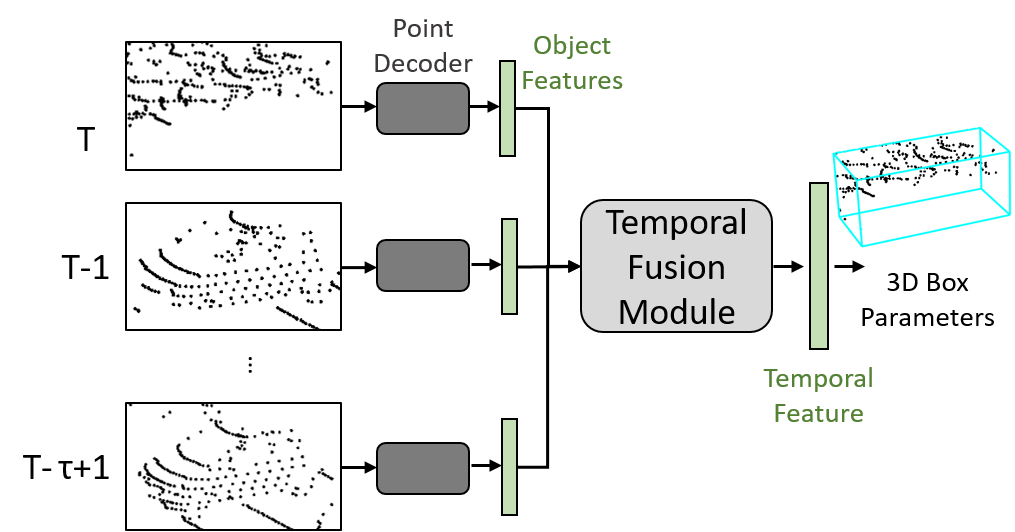}
    \caption{The proposed temporal fusion module increases 3D object detection robustness. Frame-by-frame regression is weak against noise, transient field-of-view obstruction, and sparsity. Information from past frames can be accommodated in the regression frame with temporal fusion to alleviate the aforementioned issues.}
    \label{fig:general_tfm}
\end{figure}

Using successive frames to improve the quality of the available information at a time has been extensively studied in video object detection \cite{lu2020, han2016, kang2017, ning2017} and action recognition \cite{kopuklu2019,feichtenhofer2016, wang2016, carreira2017, zolfaghari2018,jiang2019,yang2020} fields on 2D images using deep learning methods. Contrary to the width of sequence usage for 2D detection, recently proposed 3D object detection methods processing LiDAR point clouds \cite{shi2020,zhou2018,ku2018, he2020, yang20203dssd,yang2019} obtained the state of the art results with frame by frame detection. Generating 3D flow masks on point clouds has been also gaining attention recently \cite{liu2019, wang2020-flow}; however, to our best knowledge these methods have not been studied for 3D object detection. 

Instead of relying on a single frame as the previously proposed 3D object detection methods, in this study, we extend a 3D detector to use features from previous time-steps to improve 3D detection quality as shown in Fig. \ref{fig:general_tfm}. We take the Frustum PointNet architecture \cite{qi2018} as 3D object detector, which processes raw points extracted from a region of interest to predict 3D bounding boxes. The architecture generates global features per object in its 3D Amodal Box Estimation PointNet, which are used to regress the parameters as well as to predict classes. We propose a Temporal Fusion Module (TFM) to make use of the representative object-specific features in time for a better 3D object detection result. We run the network on the KITTI Multi-object Tracking Dataset \cite{geiger2012}, which provides successive frames with LiDAR data. Our methods increases the moderate difficulty AP results by ~6\%, ~4\%, and ~6\% on Car, Pedestrian, and Cyclist classes respectively comparing to the unmodified Frustum PointNet architecture that run on the same dataset. 

Our contributions are given as follows:
\begin{itemize}
    \item Frame-by-frame 3D object detection is weak against noise, field-of-view obstruction, and sparsity. We introduce a Temporal Fusion Module (TFM) to accommodate information from past frames in the regression frame to increase robustness.      
    \item According to authors best knowledge, the proposed method is the first to make use of time-sequence of object-level LiDAR features to improve 3D object detection results. 
    \item We conduct ablation studies to show how LiDAR-based features can be fused in time for 3D object detection task.
\end{itemize}

\section{RELATED WORK}
% Structure (where our research in the field)-> bold subsection
\noindent\textbf{Single- and Multi-frame 3D Object Detection.} 3D object detection relies mostly on LiDAR sensors, which are processed with methods that are based on voxelization \cite{zhou2018}, PointNets \cite{qi2018}, and projection of LiDAR data onto the Bird's eye view (BEV) \cite{ku2018}. Even though sequential data is provided in object detection datasets for autonomous driving, 3D object detectors mainly utilize a single time-step while estimating 3D bounding boxes. As an attempt to use temporal information, Yin et al. \cite{yin2020} proposed to predict a BEV velocity heatmap by targeting the center of object differences in the current and previous time-steps. However, their method utilizes only one past time-step while estimating the center. In contrast, our method provides a principled multi-frame approach to LiDAR processing for 3D object detection by fusing features from multiple previous time-steps. \cite{huang2020}, \cite{mccrae2020}, and \cite{yin2020} also propose RNN-based methods to aggregate point cloud features from multiple frames for 3D object detection. However, the methods utilize the whole point cloud, which requires motion compensation to align voxel- or pillar-based features in time. Differently, Temp-Frustum Net fuses object-level features in time to be freed from such an alignment.
\newline

\noindent\textbf{3D Object Tracking.} Temporal features have been extensively used in multi-object tracking field, especially to associate objects in successive frames. In \cite{zhang2020}, Zhang et al. learn object embeddings to find the position of the same objects in successive frames. In addition to the usual 2D detection losses, they utilize an embedding loss between the embeddings of the same objects in different frames to perform tracking. The CenterTrack algorithm \cite{zhou2020} utilizes bounding box predictions in the previous time-step as well as images from the previous and current time-steps to have a refined localization in the current frame by predicting offsets, which relates the objects of the current frame to the ones of the previous frame for tracking. Weng et al. \cite{weng2020} propose GNN3DMOT architecture to associate detected objects in successive frames by learning an affinity matrix. For this aim, they utilize appearance-based features of 2D and 3D detected object regions as well as motion features, which are the 2D and 3D bounding box parameters. The 2D and 3D appearance-based features are obtained from the image crops and LiDAR points, respectively, in the predicted regions of interest of the current and previous time-steps. Similar to these approaches, we collect the already generated 3D feature embeddings of the objects in successive frames, which are used for estimating output parameters, and combine them to obtain better 3D detection instead of tracking. 
\newline

\noindent\textbf{Multi-frame 2D Object Detection.} Sequential data have been mostly utilized for 2D video object detection with different methods. In \cite{tripathi2016}, Tripathi et al. train a recurrent network with pseudo-labels of objects in successive frames from a 2D detector to improve 2D detection scores using temporal information from videos. They propose additional losses to the regular regression and classification losses to regularize with similarity, to ensure smoothness of bounding boxes, and to avoid category changes between frames. Similarly, Lu et al. use a recurrent network to refine outputs of a 2D detector in \cite{lu2017}, however they utilize stacked tensors built from features of the detected objects as inputs to the recurrent network trained with an association loss in addition the losses used in \cite{tripathi2016}. Different from these approaches, Liu \& Zhu \cite{liu2018} utilize Bottleneck-LSTMs, a type of convolutional LSTMs, to process feature maps of the successive frames to improve 2D bounding box quality. Similar to the image-based approaches, our aim is to make use of the object-specific point features obtained in the current time-step and previous time-steps to predict better 3D bounding boxes in the current frame using LiDAR points instead of images. For this aim, we study the recurrent layers for the fusion of temporal data, which can be trained at the same time with the 3D detector. Our network differs from the previous works by studying the fusion of temporal 3D features for 3D object detection.
%\newline

%\noindent\textbf{Multi-frame 3D Object Detection.} With the recently introduced datasets \cite{caesar2020} that contain sequential data, methods to make use of multiple frames for 3D detection have been proposed. Huang et al. \cite{huang2020} utilizes LSTMs to process the voxel-based features in time in addition to motion compensation. McCrae and Zakhor \cite{mccrae2020} extend a pillar-based detector with a convolutional LSTM layer for feature propagation in time.  Yin et al. \cite{yin2020} propose spatio-temporal attention network for processing features in multiple frames to obtain better 3D detections. Even though the proposed methods provide better detection results, they all operate on the whole point cloud, which would also require motion compensation to align feature positions in time. Our method, however, investigates object-level feature fusion in time to improve 3D detection. 
\vspace{.7\baselineskip}

\section{PROPOSED METHOD}
%Problem definition
Current state-of-the-art 3D object detection methods rely on a single frame or two-frame LiDAR features to estimate objects in the environment \cite{shi2020,yin2020}. However, using the features from previous time-steps might be beneficial for better detections, especially in the case of occlusion. In this study, our aim is to show effectiveness of using multi-frame object features to improve 3D object detection. To do so, we take the v1 model of the Frustum PointNet \cite{qi2018} architecture as an object detector and extend its Amodal 3D Box Estimation Network with our Temporal Fusion Module (TFM) to fuse generated object features in time. In the following, we explain the proposed method to make use of multi-frame features. 

\begin{figure*}[ht!]
    \centering
    \includegraphics[width=0.99\textwidth]{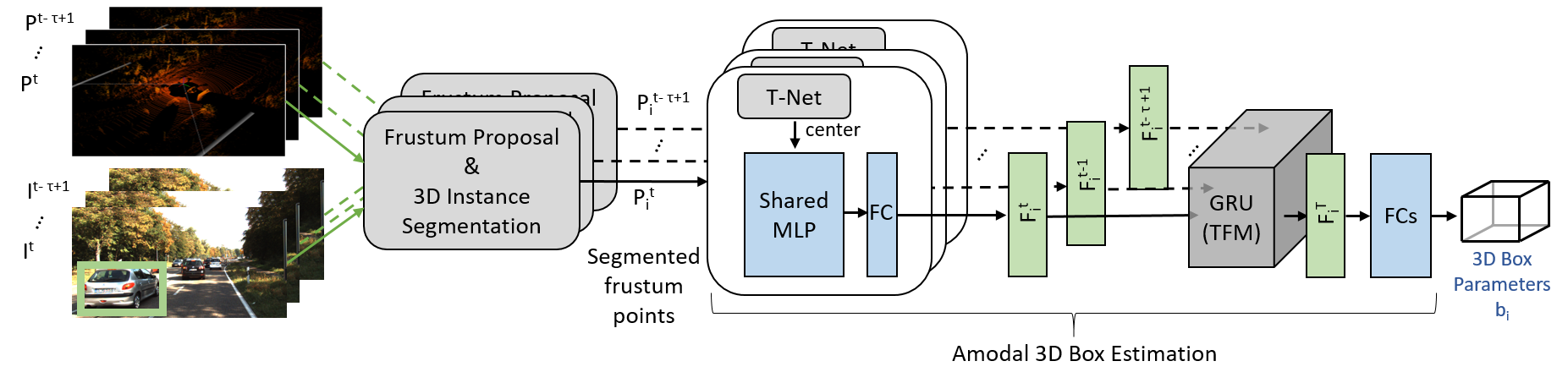}
    \caption{\textbf{3D detection network extended with the proposed Temporal Fusion Module (TFM)}: Instances of the 3D detector take LiDAR points ($P$) and 2D bounding boxes ($Q$) of images ($I$) in successive time-steps ($\{t,t-1,...,t-\tau+1\}$) and estimate the 3D bounding boxes at time-step $t$ using the generated temporal features using TFM. The Frustum Proposal and 3D Instance Segmentation networks extract the LiDAR points that belong to the frustum of the given 2D bounding box and predict a segmentation mask, respectively, to output segmented LiDAR points ($P_i$) from the frustums of objects indicated with index $i$. The segmented points and center prediction from T-Net are passed to the shared MLP and a fully-connected layer to generate a global feature ($F_i^t$) of the frustum. The $F_i^t$ and the global features of the same object from the previous $\tau-1$ time-steps ($F_i^t-1,...,F_i^t-\tau+1 $) are fed into the TFM, which consists of GRUs, to generate the temporally-enhanced feature ($F_i^T$) of the interested object, which is then passed to the fully-connected layers (FCs) to have final 3D bounding box parameters of the object ($b_i$).} %(The figure is modified and extended based on \cite{qi2018})
    \label{fig:fpnet}
\end{figure*}

\subsection{Problem Formulation}
We consider a LiDAR point cloud $P_t=\left \{ p_j| j=1, ... , n \right \}$ and an image $\textbf{I}_t \in \mathbb{R}^{H\times W \times 3}$ obtained at time-step $t$, where $n$ is the number of points, $\textbf{p}_j \in \mathbb{R}^3$ is the $j$th point in the point cloud, and $H$ and $W$ are the height and width of the image respectively. We assume that 2D bounding boxes $Q_t=\left \{ \textbf{q}_i | i=1, ... , m \right \}$ and object-specific track IDs $O_t=\left \{ o_i | i=1, ... , m \right \}$ of the objects in frame $t$ are obtained with $\textbf{q}_i \in \mathbb{R}^4$ and $o_i \in \mathbb{Z}^+$, where $m$ is the total number of objects at frame $t$.

Given the time-sequence triplets $S=\{ (P_t, Q_t, O_t), ...,\allowbreak (P_{t-\tau+1},Q_{t-\tau+1},O_{t-\tau+1}) \}$, where $\tau$ defines the sequence length in time, we are interested in finding 3D bounding boxes $B_t = \left \{ \textbf{b}_i|i=1,...,m \right \}$ in frame $t$ with $\textbf{b}_i \in \mathbb{R}^7$, which consists of size parameters ($h,w,l$), center ($c_x,c_y,c_z$), and a heading angle ($\theta$). 

\subsection{Frustum Proposal \& 3D Instance Segmentation} The Frustum PointNet detector takes the LiDAR point cloud ($P_t$) with the 2D bounding boxes ($Q_t$) to output a set of points $P^Q_t=\{P^1_t, ... , P^m_t \}$ that belongs to the object of interest given with the 2D bounding box as shown on the left-hand side of Fig. \ref{fig:fpnet}. The Frustum Proposal network extracts points in a frustum of the given 2D bounding box to limit the search space. The frustum points are directed to the 3D Instance Segmentation PointNet, where the object points are predicted by estimating a mask on the frustum points using shared multi-layer perceptrons (MLPs). The segmented points ($P^i_t$ for $i={1,...,m}$) of the frustum are then passed to the Amodal 3D Box Estimation network, where 3D bounding box parameters are estimated. For the sake of being concise, the details of Frustum Proposal and 3D Instance Segmentation parts are not explained here and readers are encouraged to see the original paper \cite{qi2018} for further details.

\subsection{Amodal 3D Box Estimation} Amodal 3D Box Estimation network takes the segmented frustum points ($P^i_t$) and outputs the final 3D bounding box parameters ($b_i$). As seen on the right-hand side of Fig. \ref{fig:fpnet}, T-Net takes the masked points and estimates the residual center of the bounding box. Building blocks of the Amodal 3D Box Estimation PointNet of the Frustum PointNet architecture is shown in Fig. \ref{fig:fpnet} with blue boxes, which generate object-specific features ($F_i^t$) and predict the 3D bounding box parameters using a shared MLP with following fully-connected layers (FCs). We place our Temporal Fusion Module (TFM) here to obtain a temporally-enhanced feature of the object of interest for more accurate detections by fusing object-specific features from multiple frames.  

\subsection{Temporal Fusion Module (TFM)} Given a set of object-specific features $F_i = \{F_i^{t}, F_i^{t-1}, ..., F^{t-\tau+1}\}$ in time, we are interested in finding a function $f$ that generates temporally-enhanced object feature $F^T_i = f(F_i)$. The proposed TFM realizes the function $f$ by processing the feature vectors of the same object from the current frame and from the previous $\tau-1$ time steps. TFM's role is to compensate the missing information in the $F_i^{t}$, which might be caused by occlusion or noise, with features $F_i^{t-1}$, $F_i^{t-2}$, ..., $F_i^{t-\tau+1}$ of the same object in time for estimating 3D box parameters of the corresponding object.

We realize the function $f$ using a recurrent layer with gated recurrent units (GRUs) \cite{cho2014} to have a time-specific feature of the object followed by a FC layer to have the final shape of the object feature. Then the time-specific object feature ($F^T_i$) is passed to the following FC layers as shown with the right-most blue box in Fig. \ref{fig:fpnet}. The $F^T_i$ is expected to be rich in representation, which would result in better detection scores than the baseline, the original Frustum PointNet detector. We also concatenate the center residual prediction of the same object from T-Net with the object features to have more representative features, which are also evaluated in \ref{subsec:results}.

\subsection{Output Branching} The TFM can be placed differently in the Amodal 3D Box Estimation network. The straightforward approach is shown in Fig. \ref{fig:fpnet} using the output of TFM ($F^T_i$) directly to predict all of the box parameters.  We call this setup as One Branch (OB) in our ablation studies (\ref{subsec:ablation}). However, even though the objects from subsequent frames are the same, their parameters differ between frames depending on the sampling rates. Thus, the features must also contain information specific to the position. From this point of view, we also tested having two branches and placing the TFM only to one of the branches. The feature of the first branch is taken from the current time-step and the feature of the second branch is generated temporally through TFM. Finally, the predicted parameters of the two branches are averaged to get the final bounding box. This is called as Two Branch (TB) version in our ablation studies. 

The box parameters consist of centers, orientation and confidence scores of the orientation bins, and size regressions. Since most of the traffic participants are moving, the position of the object and points in the frustum changes. Therefore, only the size parameters are shared between successive frames and the rest of the parameters will have a shift depending on the sampling frequency of the data. Also taking this into account, we decided to use temporal features only to predict size parameters in the Two Branch version. Namely, the second branch with TFM is used to predict the size residuals and the first branch is used for estimating rest of the parameters (Ours).

%%%%%%%%%%%%%%%%%%%%%%%%%%%%%%%%%%%%%%%%%%%%%%%%%%%%%%%%
%%%%%%%%%%%%%%%%%%%%%%%%%%%%%%%%%%%%%%%%%%%%%%%%%%%%%%%%
%%%%%%%%%%%%% EXPERIMENTS AND RESULTS %%%%%%%%%%%%%%%%%%
%%%%%%%%%%%%%%%%%%%%%%%%%%%%%%%%%%%%%%%%%%%%%%%%%%%%%%%%
%%%%%%%%%%%%%%%%%%%%%%%%%%%%%%%%%%%%%%%%%%%%%%%%%%%%%%%%

\section{EXPERIMENTS \& RESULTS}

\subsection{Dataset}
\label{subsec:dataset}
\begin{figure}[ht!]
    \begin{subfigure}[b]{.5\linewidth}
        \centering
        \includegraphics[width=0.9\textwidth]{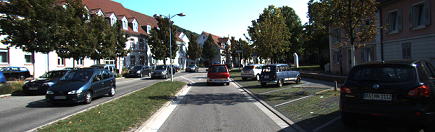}
        \caption{Drive 11}
    \end{subfigure}%
    \begin{subfigure}[b]{.5\linewidth}
        \centering
        \includegraphics[width=0.9\textwidth]{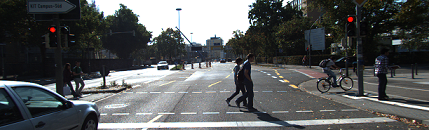}
        \caption{Drive 15}
    \end{subfigure}

    \begin{subfigure}[b]{.5\linewidth}
        \centering
        \includegraphics[width=0.9\textwidth]{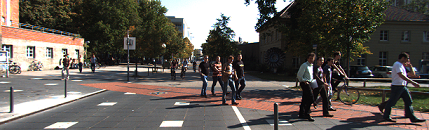}
        \caption{Drive 16}
    \end{subfigure}%
    \begin{subfigure}[b]{.5\linewidth}
        \centering
        \includegraphics[width=0.9\textwidth]{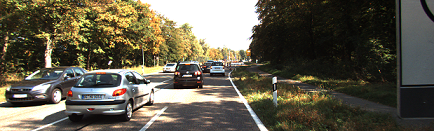}
        \caption{Drive 18}
    \end{subfigure}
    \caption{Scenes from KITTI Object Tracking Dataset drives that are used for validation.}\label{fig:dataset}
    
\end{figure}

The KITTI object tracking benchmark \cite{geiger2012} consists of 21 training drives, whose labels contain the ground-truth track IDs with 3D bounding boxes. All the drives consist of camera images and LiDAR point clouds. Therefore, the dataset is suitable for using feature vectors of the same object from successive frames. We used drives 11, 15, 16, and 18 for validation of the network and the rest of the drives for training to balance the distribution of the class members in the training and validation sets similarly. There are 6264 frames, 31886 car, 8378 pedestrian, and 1039 cyclist instances in the training set and 1239 frames, 6494 car, 2980 pedestrian, and 809 cyclist instances in the validation set. The validation drives are selected to be challenging and example scenes can be seen in Fig. \ref{fig:dataset}. The drive 11 and 15 are urban drives, denser comparing to others. In drive 15, pedestrians crossing the street occlude the far away cars, which are already represented with a small number of points in the non-occlusion case. In drive 16, the ego-vehicle doesn't move and pedestrians cross in front of the parked cars behind, which causes invisible parts of the cars to the ego-vehicle. Drive 18 is an outer city drive, in which the ego-vehicle approaches the traffic and there is only minor changes in the occlusion. 

%[thpb]

\subsection{Loss function}
\label{subsec:loss}
We keep the original loss function of the Frustum PointNet architecture, where Huber loss is used for the regression of centers, box sizes, and the orientation angle. The cross-entropy loss is used for classification scores, segmentation of points, and orientation angle bin scores. Additional to these, we also test the cosine distance loss to provide the feature embeddings of the same object to be similar in successive frames. The cosine distance loss is calculated by taking the cosine difference between two feature vectors of the same object in subsequent frames as given with the Eq. \ref{eq:cos_dist}. In the equation $\upsilon$ and $\omega$ are the feature vectors, with $l$ dimensions, of the object from successive frames. 

\begin{equation}\label{eq:cos_dist}
    \begin{aligned}
    cosine\, distance(\upsilon, \omega ) = 1 - \frac{\sum_{k=1}^{l} \upsilon_k \omega_k}{\sqrt{\sum_{k=1}^{l}\upsilon_k^2} \sqrt{\sum_{k=1}^{l}\omega_k^2}}
    \end{aligned}
\end{equation}

% https://www.tensorflow.org/api_docs/python/tf/compat/v1/losses/cosine_distance

\subsection{Technical Details \& Training}
The shared MLP shown in Fig. \ref{fig:fpnet} consists of (128,128,256,512)  unit  layers. The FCs, which are shown after TFM, have 512 and 256 units. In the experiments, the TFM is placed between these two layers to get the features from the output of the former FC layer. A final FC layer is used to estimate box parameters. We used a 512-unit GRU in the TFM followed by a 512-unit FC layer. The point features are extracted only based-on x, y, z values without reflection value.

The network is trained with a batch size of 32 through 100 epochs using Tensorflow 1.15. The number of points sampled in a frustum is set to 1024. The network parameters are optimized with Adam optimizer with a momentum value 0.9 and a learning rate of 0.001. Other parameters are experimented (as explained in \ref{subsec:experiments}) and the results are provided in \ref{subsec:results}. Since this study aims to show the usability of feature vectors from the successive frames to compensate loss of information, we utilized ground-truth 2D bounding box information while extracting frustums and also tracking IDs for matching object feature vectors in successive frames. For all the networks shown in \ref{subsec:experiments}, we used the same ground-truths to be sure that the improvement comes only from the temporal fusion. 

\begin{figure*}[ht!]
\centering
    \begin{subfigure}[b]{0.98\textwidth}
        \centering
        \includegraphics[width=0.98\textwidth]{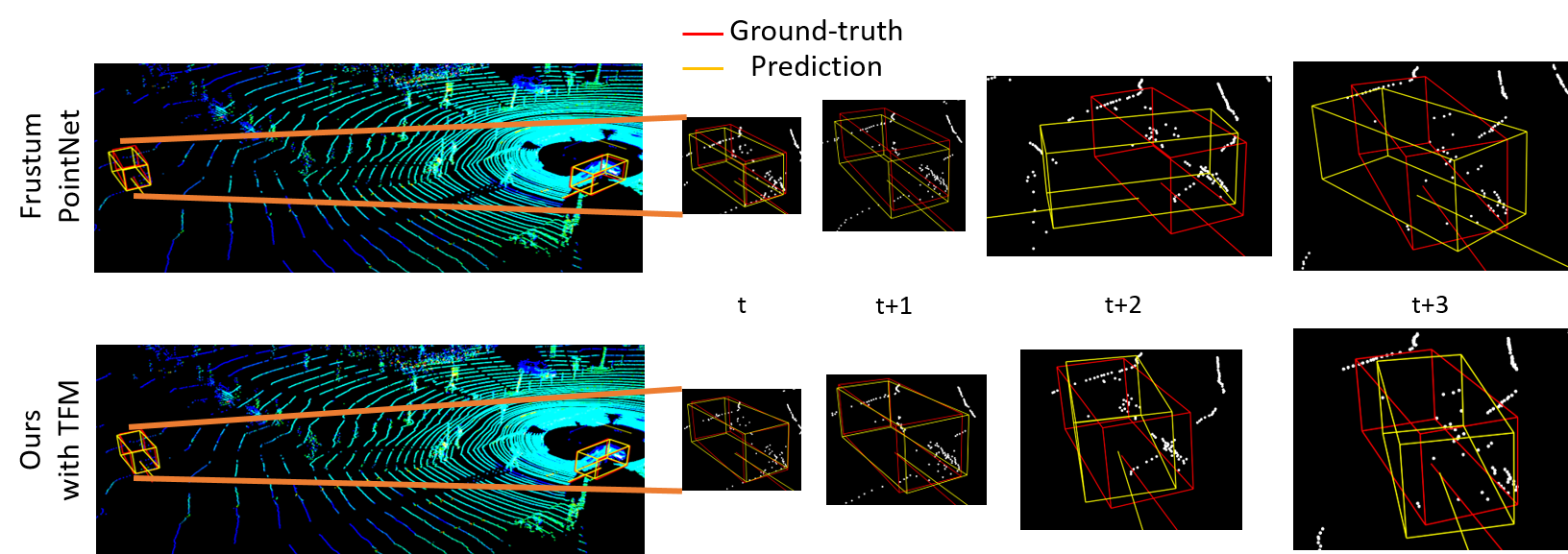}
        \caption{Car detection results on KITTI Tracking Dataset Drive 15}
    \end{subfigure} 
    \\
    \begin{subfigure}[b]{0.98\textwidth}
        \centering
        \includegraphics[width=0.98\textwidth]{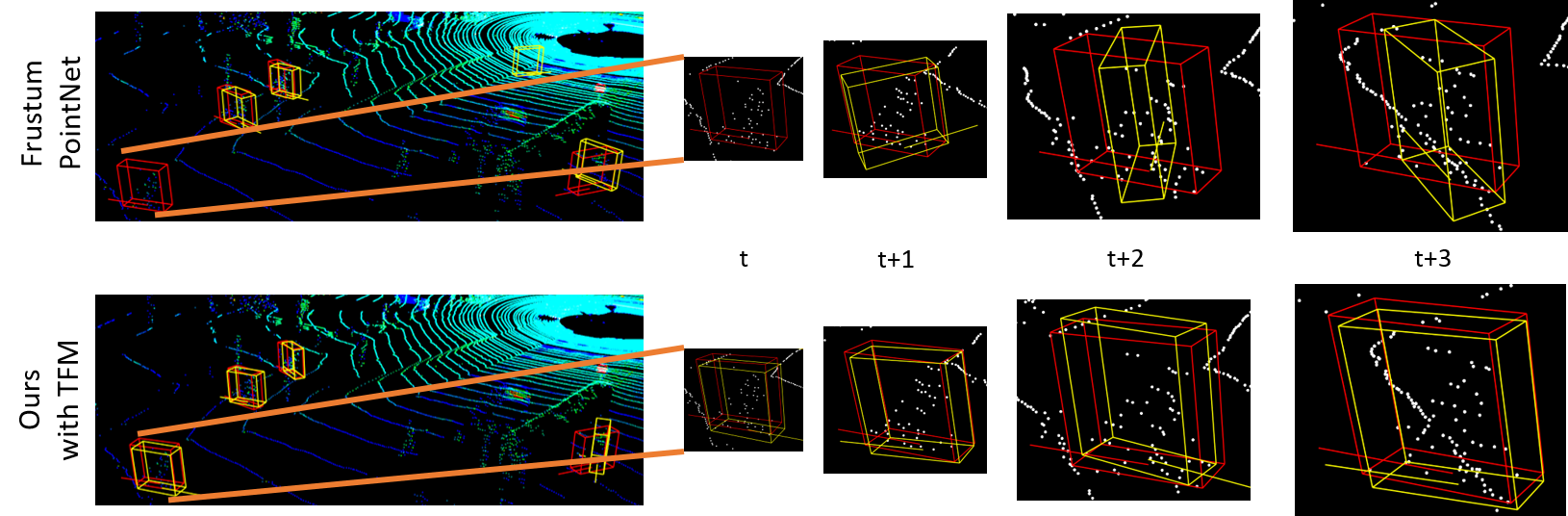}
        \caption{Cyclist detection results on KITTI Tracking Dataset Drive 15}
    \end{subfigure}
    
    \caption{\textbf{Comparison of the qualitative 3D detection results} on a frame sequence in drive 15 of KITTI tracking dataset for Car and Cyclist detection (Red: Ground-truth, Yellow: Predictions). The detections at the top of (a) and (b) are obtained with the unmodified Frustum PointNet \cite{qi2018} and the ones below them are obtained with the proposed method (Ours). Even though Frustum PointNet detected the Car on the left in the first two frames in (a), the 3D bounding boxes for the following two frames were not accurate. In contrast, our proposed method could keep accurate detections while pedestrians were crossing the street in front of the ego-vehicle, which occludes the far-away Car (shown in the zoomed-in images). Similarly, bounding boxes of Cyclists detected with Frustum PointNet fluctuate between frames, however, the proposed method provides more consistent bounding boxes in (b) at the bottom.}
    \label{fig:qualitative}
\end{figure*}

\subsection{Experiments}
\label{subsec:experiments}

We evaluate the effectiveness of the proposed method by comparing it with the results obtained by unmodified Frustum PointNet architecture on the KITTI tracking dataset as described in \ref{subsec:dataset} for Car, Pedestrian, and Cyclist classes. We use the Average Precision (AP) metric, which is calculated in the same way as calculated for the KITTI Object Detection Benchmark. The AP scores are obtained with Intersection over Union (IoU) thresholds of $0.7$, $0.5$, and $0.5$ for Car, Pedestrian, and Cyclist classes respectively. Additionally, we evaluate the performance of the TFM with the cosine distance loss. Results are given in \ref{subsec:results}. 

We also run ablation studies (\ref{subsec:ablation}) to show the effectiveness of the selected placement of the TFM inside the detector. Also, we evaluate further ablation experiments to show that the improvement comes from the temporal fusion, but not from the additional depth brought by the TFM. 

The AP results are given for the difficulty levels of the KITTI dataset as given by Easy, Moderate, and Hard difficulties depending on the 2D bounding box size, occlusion, and truncation. For each row of the tables, we run the training at least 5 times and took the best results according to the moderate difficulty.

\subsection{Results}
\label{subsec:results}

In Table \ref{tab:results}, the proposed method (Ours) outperforms the baseline Frustum PointNet detector for Pedestrian and Cyclist classes in all difficulty levels and for the moderate difficulty of Car class. We also evaluated the TFM when the predicted centers of the objects from T-Net are added to the temporal features, which is given by "Ours(w/ C)" in the table. This version only achieved better results than "Ours" for Cyclist class. $\tau$ column shows the number of time-steps, for which the best scores were obtained. We changed $\tau$ between $2$ and $6$ and for most of the versions we already obtained the best results with $\tau=3$ or $\tau=4$. Adding cosine distance loss (Table \ref{tab:results} on the right-hand side) further improved the best results for Pedestrian and Cyclist classes by $~3\%$ and $~2\%$ in moderate difficulty, however, that didn't improve for Car class. As seen from Table \ref{tab:results}, the most improvements are obtained for moderate and hard level difficulties. This result is also aligned with our expectations. Thinking of a track of an object, the temporally-enhanced features are most effective when the LiDAR returns a small number of points from the object. An object mostly falls into the easy category when it is closest to the ego-vehicle, which is either the start of a track or the end of a track due to the approaching or driving away behavior of the objects or the ego-vehicle. In both cases, the temporal features would bring less additional information. Therefore, the improvement seen in moderate and hard levels are proportionally more than the easy level, however improvements are still comparable. 

All in all, we obtain $~6\%$, $~4\%$, and $~6\%$ improvement in the moderate AP using the TFM method without cosine distance loss on Car, Pedestrian, and Cyclist classes respectively. The qualitative results can be seen in Fig. \ref{fig:qualitative}, which shows the detection comparison on frames of a sequence obtained by Frustum PointNet and the proposed method (Ours). We see that the detections of Frustum PointNet in the first two frames could not be preserved in the following two frames. Oppositely, the proposed method could still provide accurate detections in the sequence. The qualitative and quantitative results also suggest that the loss used for size estimation on the temporal features take a role of an auxiliary loss for the other parameters, since two branches in (Ours) share a root feature, which helped estimating other parameters more accurately. 

\begin{table}[t]%[h]
\vspace{0.5cm}
 \caption{\textbf{The best AP results} (IoU=0.7, IoU=0.5, and IoU=0.5) for Car, Pedestrian, and Cyclist classes on KITTI tracking validation drives}
 \label{tab:results}
 \begin{center}
\begin{tabular}{ccccc|cccc|}
\cline{6-9}
                                             &        &      &      &      & \multicolumn{4}{c|}{Cosine Loss} \\ \hline
\multicolumn{1}{|c|}{\textbf{Car}}                    & $\tau$ & Easy & Mod & Hard & $\tau$   & Easy  & Mod  & Hard  \\ \hline
\multicolumn{1}{|c|}{\begin{tabular}[c]{@{}c@{}}Frustum\\ PointNet\cite{qi2018}\end{tabular}}               & -      & 83.1 & 65.4 & \textbf{65.1} & -        & -     & -     & -     \\ \hline
\multicolumn{1}{|c|}{Ours} & 4      & 82.5 & \textbf{71.8} & 64.8   & 4        & 82.5  & 65.5  & 64.1  \\
\multicolumn{1}{|c|}{Ours(w/ C)}       & 4      & \textbf{83.7} & 65.6 & 64.1 & 3        & 81.2  & \textbf{65.6}    & 63.9  \\ \hline \hline

\multicolumn{1}{|c|}{\textbf{Pedestrian}} & $\tau$ & Easy & Mod  & Hard & $\tau$   & Easy  & Mod   & Hard  \\ \hline
\multicolumn{1}{|c|}{\begin{tabular}[c]{@{}c@{}}Frustum\\ PointNet\cite{qi2018}\end{tabular}}   & -      & 63.6 & 56.5 & 50.4 & -        & -     & -     & -     \\ \hline
\multicolumn{1}{|c|}{Ours}      & 3      & \textbf{67.5} & \textbf{60.4} & \textbf{53.6} & 4        & 65.7  & \textbf{63.3}  & \textbf{56.4}  \\
\multicolumn{1}{|c|}{Ours(w/ C)}    & 3      & 64.5 & 59.4 & 52.7 & 4        & 65.6  & 59.1  & 52.6  \\ \hline \hline

\multicolumn{1}{|c|}{\textbf{Cyclist}}  & $\tau$ & Easy & Mod  & Hard & $\tau$   & Easy  & Mod   & Hard  \\ \hline
\multicolumn{1}{|c|}{\begin{tabular}[c]{@{}c@{}}Frustum\\ PointNet\cite{qi2018}\end{tabular}} & -      & 81.3 & 59.4 & 58.8 & -        & -     & -     & -     \\ \hline
\multicolumn{1}{|c|}{Ours}    & 3      & 84.9 & 65.3 & 65   & 3        & 85.6  & 65.9  & 65.3   \\
\multicolumn{1}{|c|}{Ours(w/ C)}  & 4      & \textbf{86.2} & \textbf{74.5} & \textbf{67.3}   & 6        & \textbf{87.5}  & \textbf{76.8}  & \textbf{69.7}  \\ \hline

\end{tabular}
\end{center}
\end{table}

\subsection{Ablation Study}
\label{subsec:ablation}

Table \ref{tab:ablation_branch} shows the comparison between different placements of TFM in the Amodal 3D Box Estimation Network. OB stands for one branch case, in which the TFM output is used directly to estimate the 3D box parameters. TB stands for two branch case, in which two branches are generated to estimate all the box parameters. The first branch uses the feature vector of the current time-step to estimate box parameters, whereas the second branch uses TFM output for box parameter estimation. The estimated parameters are averaged at the output to obtain the final detection. "Ours" is also a two branch case, but the TFM branch is only used to estimate size parameters. Shown in Table \ref{tab:ablation_branch}, the selected placement (Ours) provides the best results comparing to the others with and without cosine loss training. The results are also aligned with our expectations. Depending on the sampling frequency of the data, there is a shift in the center and orientation of the objects due to the object and ego-vehicle movement. However, the size remains the same, which makes it beneficial to estimate the parameters using temporally-enhanced features. 

\begin{table}[t]%[h]
\vspace{0.5cm}
 \caption{\textbf{Ablation of the TFM placement:} AP (IoU=0.7, IoU=0.5, and IoU=0.5) results for Car, Pedestrian, and Cyclist classes on KITTI validation drives}
 \label{tab:ablation_branch}
 \begin{center}
\begin{tabular}{ccccc|cccc|}
\cline{6-9}
                                             &        &      &      &      & \multicolumn{4}{c|}{Cosine Loss} \\ \hline
\multicolumn{1}{|c|}{\textbf{Car}}                    & $\tau$ & Easy & Mod & Hard & $\tau$   & Easy  & Mod  & Hard  \\ \hline
\multicolumn{1}{|c|}{Ours(OB)}             & 3      & 79.9 & 61.7 & 60.2 & 3        & 79.3  & 62.1  & 61.1  \\
\multicolumn{1}{|c|}{Ours(TB)}             & 4      & 82.2 & 62.5 & 61.7 & 3        & 81.5  & 63.8  & 62.6  \\
\multicolumn{1}{|c|}{Ours} & 4      & \textbf{82.5} & \textbf{71.8} & \textbf{64.8}   & 4        & \textbf{82.5}  & \textbf{65.5}  & \textbf{64.1}  \\ \hline \hline

\multicolumn{1}{|c|}{\textbf{Pedestrian}} & $\tau$ & Easy & Mod  & Hard & $\tau$   & Easy  & Mod   & Hard  \\ \hline
\multicolumn{1}{|c|}{Ours(OB)}         & 6      & 57.8 & 51.3 & 49.8 & 6        & 66    & 59.1  & 52.6  \\
\multicolumn{1}{|c|}{Ours(TB)}         & 4      & 57.7 & 51.8 & 50.7 & 6        & \textbf{68}    & 60.6  & 54    \\
\multicolumn{1}{|c|}{Ours}      & 3      & \textbf{67.5} & \textbf{60.4} & \textbf{53.6} & 4        & 65.7  & \textbf{63.3}  & \textbf{56.4}  \\ \hline \hline

\multicolumn{1}{|c|}{\textbf{Cyclist}}  & $\tau$ & Easy & Mod  & Hard & $\tau$   & Easy  & Mod   & Hard  \\ \hline
\multicolumn{1}{|c|}{Ours(OB)}       & 4      & 77.5 & 50.9 & 50.8 & 3        & 79.5  & 64.4  & 64    \\
\multicolumn{1}{|c|}{Ours(TB)}       & 4      & 80.5 & 63.7 & 63.1 & 3        & 82.3  & 65.3  & 64.5  \\
\multicolumn{1}{|c|}{Ours}    & 3      & \textbf{84.9} & \textbf{65.3} & \textbf{65}   & 3        & \textbf{85.6}  & \textbf{65.9}  & \textbf{65.3}   \\ \hline

\end{tabular}
\end{center}
\end{table}

Adding TFM increases the network's depth, which would also improve the detection results independent from the fusion of features temporally. Therefore, Table \ref{tab:ablation_tau} indicates the comparison between the results of the proposed method from Table \ref{tab:results} (without cosine loss) and the results when $\tau$ is set to $1$. Setting $\tau$ to $1$ means using the weights of the additional layers without enhancing the object features temporally. As seen from Table \ref{tab:ablation_tau}, $\tau>1$ cases provide better results than the network with $\tau=1$ showing the usefulness of temporal fusion. 

\begin{table}[t]%[h]
\caption{Comparison of $\tau=1$ with the best $\tau$ for Car, Pedestrian, and Cyclist classes}
\label{tab:ablation_tau}
\begin{center}
\begin{tabular}{|c|cccc|}
\hline
\textbf{Car}     & $\tau$ & Easy & Mod  & Hard \\ \hline
Ours   & 1      & 84.1 & 64.6 & 63.4 \\
Ours   & 4      & 82.5 & 71.8 & 64.8   \\
Ours(w/ C) & 1      & 82.9 & 63.9 & 62.4 \\
Ours(w/ C) & 4      & 83.7 & 65.6 & 64.1 \\ \hline
\hline
\textbf{Pedestrian} & $\tau$ & Easy & Mod  & Hard \\ \hline
Ours      & 1      & 61.7 & 50.9 & 49.8 \\
Ours      & 3      & 67.5 & 60.4 & 53.6 \\
Ours(w/ C)    & 1      & 64   & 57.9 & 51.8 \\
Ours(w/ C)    & 3      & 64.5 & 59.4 & 52.7 \\ \hline
\hline
\textbf{Cyclist} & $\tau$ & Easy & Mod  & Hard \\ \hline
Ours   & 1      & 86.5 & 65   & 64.6 \\
Ours   & 3      & 84.9 & 65.3 & 65   \\
Ours(w/ C) & 1      & 85.3 & 67.6 & 66.9 \\
Ours(w/ C) & 4      & 86.2 & 74.5 & 67.3  \\ \hline

\end{tabular}
\end{center}
\end{table}

\section{CONCLUSION}
In this study, we extend the Frustum PointNet architecture with the proposed Temporal Fusion Module (TFM) to use the features from the previous time-steps for having better 3D detections in the current frame. The TFM takes the feature vectors of the same object from the current frame and the previous frames and learns a more representative feature vector for predicting 3D bounding box parameters. We experimented different branching versions on the KITTI tracking dataset for placing the TFM in the original process. We show that fusing features of an object in time with our TFM provides better AP results. This holds for moderate level difficulty for Car, Pedestrian, and Cyclist classes by improving the baseline with ~6\%, ~4\%, and ~6\% in AP. Additionally, the training with cosine distance loss yielded even higher AP values for Pedestrian and Cyclist classes, however did not improve Car class detections. All in all, we show the efficiency of using temporal features for improving 3D object detection as a proof-of-concept on the Frustum PointNet architecture. We find the results promising for the future work, in which the proposed method is worth to be implemented on more recent state-of-the-art detectors to improve the 3D detection scores.

\addtolength{\textheight}{-12cm}   % This command serves to balance the column lengths
                                  % on the last page of the document manually. It shortens
                                  % the textheight of the last page by a suitable amount.
                                  % This command does not take effect until the next page
                                  % so it should come on the page before the last. Make
                                  % sure that you do not shorten the textheight too much.
                                  % was -12cm 

%%%%%%%%%%%%%%%%%%%%%%%%%%%%%%%%%%%%%%%%%%%%%%%%%%%%%%%%%%%%%%%%%%%%%%%%%%%%%%%%

%%%%%%%%%%%%%%%%%%%%%%%%%%%%%%%%%%%%%%%%%%%%%%%%%%%%%%%%%%%%%%%%%%%%%%%%%%%%%%%%

%%%%%%%%%%%%%%%%%%%%%%%%%%%%%%%%%%%%%%%%%%%%%%%%%%%%%%%%%%%%%%%%%%%%%%%%%%%%%%%%
%\section*{APPENDIX}

%Appendixes should appear before the acknowledgment.

\section*{ACKNOWLEDGMENT}
This work was funded in part by the United States Department of Transportation
under award number 69A3551747111 for Mobility21: the National University
Transportation Center for Improving Mobility.

Any findings, conclusions, or recommendations expressed herein are those
of the authors and do not necessarily reflect the views of the United
States Department of Transportation, Carnegie Mellon University, or The Ohio State University.

%%%%%%%%%%%%%%%%%%%%%%%%%%%%%%%%%%%%%%%%%%%%%%%%%%%%%%%%%%%%%%%%%%%%%%%%%%%%%%%%

\end{document}